\documentclass[letterpaper, 10 pt, conference]{ieeeconf}  

\IEEEoverridecommandlockouts                              

\usepackage{graphics} 
\usepackage{graphicx}
\usepackage{amsmath} 
\usepackage{amssymb}  
\usepackage{multirow}
\usepackage{subfigure}
\usepackage{floatrow}
\usepackage{array}
\usepackage{cite}
\usepackage{multicol}
\usepackage{multirow}
\usepackage{makecell}
\usepackage{epstopdf}
\usepackage{bbding}
\usepackage{bbm}
\usepackage{bm}
\usepackage{url}            
\usepackage{booktabs}       
\usepackage{amsfonts}       
\usepackage{nicefrac}       
\usepackage[pagebackref,breaklinks,colorlinks]{hyperref}
\usepackage{microtype}      

\usepackage{pifont}
\newcommand{\cmark}{\ding{51}}

\DeclareMathOperator*{\argmin}{arg\,min}

\usepackage[ruled,linesnumbered]{algorithm2e}
\SetKwComment{Comment}{/* }{ */}

\title{\LARGE \bf
Pyramid Semantic Graph-based Global Point Cloud 
 Registration\\ with Low Overlap
}

\author{
Zhijian Qiao, Zehuan Yu, Huan Yin and Shaojie Shen
\thanks{This work was supported in part by the HKUST Postgraduate Studentship, in part by the HKUST-DJI Joint Innovation Laboratory, and in part by the Hong Kong Center for Construction Robotics (InnoHK center supported by Hong Kong ITC).}
\thanks{The authors are with the Department of Electronic and Computer Engineering, The Hong Kong University of Science and Technology, Hong Kong, China. E-mail: zqiaoac@connect.ust.hk, zyuay@connect.ust.hk, eehyin@ust.hk, eeshaojie@ust.hk}
\thanks{Corresponding author: Huan Yin}
}

\begin{document}
\maketitle
\thispagestyle{empty}
\pagestyle{empty}

\begin{abstract}
Global point cloud registration is essential in many robotics tasks like loop closing and relocalization. Unfortunately, the registration often suffers from the low overlap between point clouds, a frequent occurrence in practical applications due to occlusion and viewpoint change. In this paper, we propose a graph-theoretic framework to address the problem of global point cloud registration with low overlap. To this end, we construct a consistency graph to facilitate robust data association and employ graduated non-convexity (GNC) for reliable pose estimation, following the state-of-the-art (SoTA) methods. 

Unlike previous approaches, we use semantic cues to scale down the dense point clouds, thus reducing the problem size. Moreover, we address the ambiguity arising from the consistency threshold by constructing a pyramid graph with multi-level consistency thresholds. Then we propose a cascaded gradient ascend method to solve the resulting densest clique problem and obtain multiple pose candidates for every consistency threshold. Finally, fast geometric verification is employed to select the optimal estimation from multiple pose candidates. Our experiments, conducted on a self-collected indoor dataset and the public KITTI dataset, demonstrate that our method achieves the highest success rate despite the low overlap of point clouds and low semantic quality. We have open-sourced our code\footnote{\href{https://github.com/HKUST-Aerial-Robotics/Pagor}{https://github.com/HKUST-Aerial-Robotics/Pagor}} for this project.
\end{abstract} 

\section{INTRODUCTION}
\label{sec:introduction}

Global point cloud registration refers to registering target and source point clouds without any initial guess. As a fundamental problem, it is critical in re-localization and loop closing for mobile robots, and object pose estimation for robotic grasping. Achieving robust and accurate global point cloud registration is a challenging task due to two main problems. First, the a priori poses are unknown, leading to a large search space for the solution. Second, the overlap between the point cloud pair is often low in practical applications due to viewpoint change or occlusion, generating many outliers.

Most state-of-the-art (SoTA) methods address these problems by employing a graph-theoretic framework to obtain robust data associations and solve for robust pose estimation~\cite{yang2020teaser,lim2022single,yin2023segregator}. More concretely, a consistency graph is first constructed, where the vertices are all correspondences obtained by feature matching or the "all to all" strategy. The edges represent the consistency between two correspondences, where the edges can be binary or weighted. A threshold is then chosen to distinguish between correspondence inliers and outliers, and guarantee the graph's sparsity. Subsequently, a maximum or densest clique algorithm is applied to obtain the set of true correspondences, based on the common assumption that, at some consistency threshold, the graph consisting of the true correspondences is the largest or edge-weighted clique. Finally, based on these correspondences, the robust pose estimation problem is modeled as a truncated least squares problem to be solved.

While these methods can resist high outlier rates and achieve robust global registration, they can still face three possible problems, given as follows. 1) The choice of the consistency threshold usually depends on the noise distribution (e.g., the "3-sigma" noise bound), which can be difficult to estimate accurately, especially when the correspondences are obtained by feature matching. 2) The assumption of maximum or densest clique may not hold in cases of unsuitable consistency threshold and low overlap rate. Outliers of non-overlapping regions in practical applications may have similar or reasonable structures with overlapping regions. 3) The problem's scale could become very large due to the availability of dense point clouds for most range sensors, leading to an unacceptable computation time.

\begin{figure}[tbp]
	\centering
	\includegraphics[width=\linewidth]{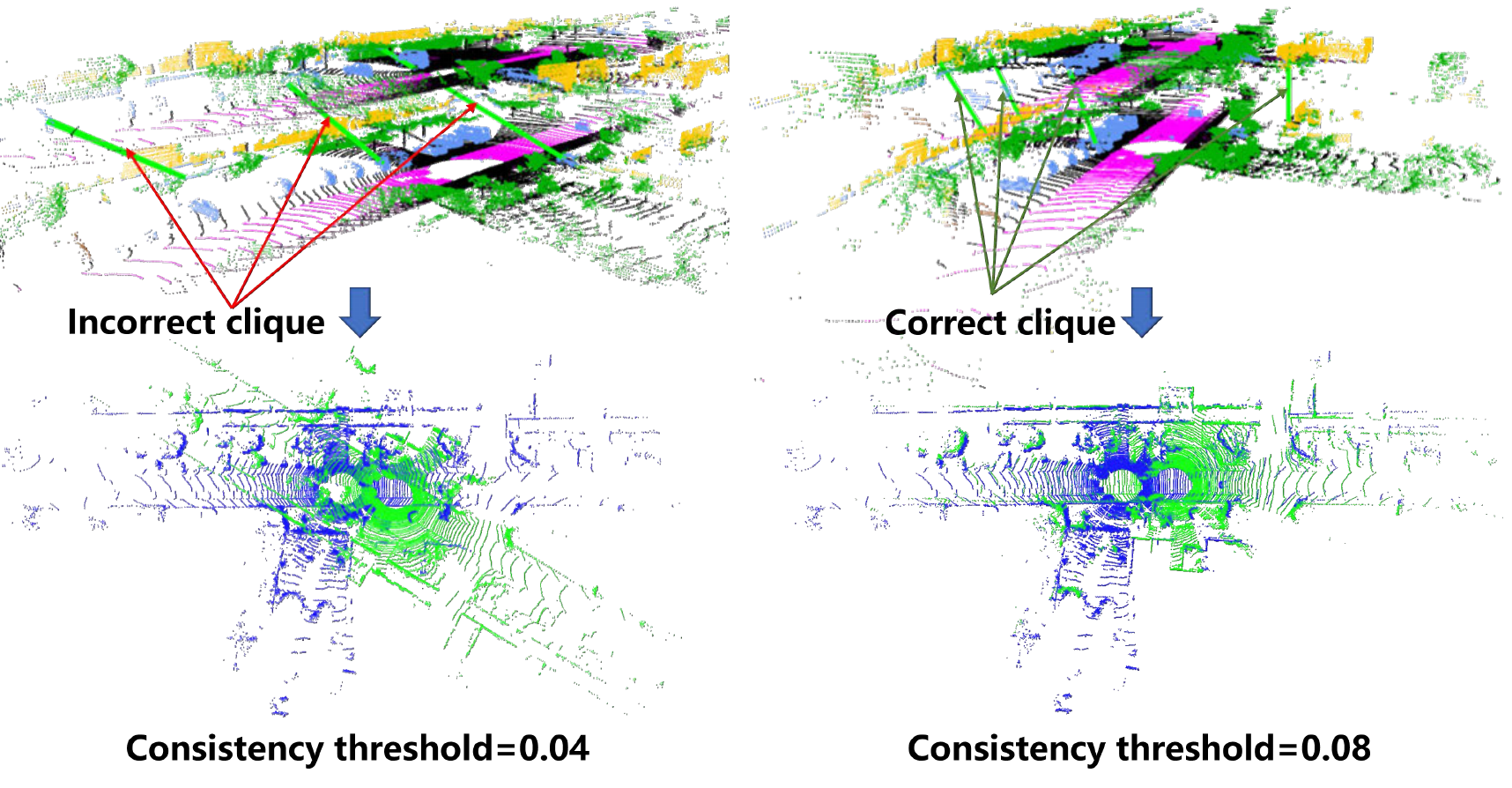}
	\caption{Visualization of correspondences (data association) and registration between two point clouds with low overlap. The difference between the left and right sides lies in the consistency threshold to construct a consistency graph. A maximum clique solved at an unreasonable threshold can produce incorrect correspondences and fail the registration. }
	\label{figure:cover}
\end{figure}

In this study, we propose a novel ``distrust and verify" method that integrates high-level semantic and low-level geometric cues. Specifically, our proposed method first builds a pyramid semantic consistency graph, in which the pyramid is achieved by shrinking the consistency thresholds on semantic landmarks. This means that we do not believe that, there exists a perfect consistency threshold that can handle any global point cloud registration cases. Additionally, the use of semantics significantly reduces the problem size and enhances the accuracy of the data association. We then employ a cascaded gradient ascent method to obtain multiple data association candidates from the pyramid consistency graph. Since inaccurate and sparse semantic landmarks may lead to degeneration, pose candidates are derived using the geometric nonlinear optimization method graduated non-convexity (GNC)~\cite{yang2020graduated} and subsequently checked through a geometric verification using low-level dense metric points. In summary, our contributions are as follows:

\begin{itemize}
\item We present a pyramid semantic graph-based global point cloud registration (Pagor) with low overlap, following a distrust and verify strategy to register using semantic cues and verify using geometric cues.
\item To handle the consistency-threshold ambiguity, we design a semantic pyramid graph with multi-level consistency thresholds and then propose a cascaded gradient ascend method to obtain multiple transformation candidates for each consistency threshold.
\item In the experimental section, we validate the proposed method in two challenging situations: re-localizing a robot globally on an indoor floor plan map with sparse laser scans; global registration with low overlap on the publicly available KITTI dataset.
\end{itemize}

\section{RELATED WORK}
\label{sec:related}

\subsection{Global Point Cloud Registration}

Generally, 3D point clouds from LiDAR are textureless and in an irregular format. These properties bring difficulties for global registration. There are mainly two pipelines for global point cloud registration: correspondence-free~\cite{bernreiter2021phaser,zhu2022correspondence} and correspondence-based methods. In this study, our proposed method relies on the correspondences generated from semantics, and we present a brief review of correspondence-based global registration.

One of the well-known registration methods is Iterative Closest Point (ICP)~\cite{besl1992method}. ICP estimates the correspondences and relative transformation in an expectation-maximization manner. ICP and its variants easily fall into local minima for global registration tasks. To address this problem, researchers proposed to extract handcrafted features on points, like FLIRT~\cite{tipaldi2010flirt} for 2D laser scans and FPFH~\cite{rusu2009fast} for 3D point clouds. With the popularity of deep learning, learning-based point cloud registration~\cite{wang2019deep,yew2022regtr} has drawn much attention in recent years. These data-driven methods generally rely on training data to guarantee generalization ability. If the correspondences of local features are known, e.g., building correspondences with FPFH~\cite{rusu2009fast}, the transformation estimation can be solved in a closed form~\cite{horn1987closed,arun1987least}. But conventional transformation solvers are easily affected by outliers in correspondences, resulting in inaccurate transformation estimation. Recent works proposed to use graph-theoretic frameworks to estimate correct correspondences~\cite{yang2020teaser,lusk2021clipper}. Furthermore, a robust kernel function is also an alternative choice to overcome outliers in transformation estimation, like graduated non-convexity~\cite{yang2020graduated}.

In this study, our proposed method is partially based on previous works~\cite{yang2020teaser,lusk2021clipper}: extract invariant measurements, and check the consistency of correspondences, and followed by robust estimation. But differently, we propose to use multi-level consistency thresholds and propose to use geometric properties to verify results, making the graph-theoretic framework more robust in challenging situations.

\subsection{Semantic-aided Robot Localization and Mapping}

Semantic information is gradually integrated into recent point cloud-based robotic tasks, particularly for robot localization and mapping. The raise of semantics is thanks to the development of front-end perception techniques. Semantic ICP~\cite{parkison2018semantic} is formulated in an expectation-maximization manner to reduce alignment errors. For indoor localization, floor plans and building information modeling can provide semantic landmarks as assistance. Hendrikx et al.~\cite{hendrikx2021connecting} proposed to build data associations between building elements and observed semantics, and the associations are used to localize a wheeled robot. Semantic building information can also be integrated into a classical ICP scheme~\cite{yin2023semantic}. In~\cite{zimmerman2022long}, lifelong visual localization is conducted on floor plan maps which can provide stable semantic cues in indoor scenes. As for outdoor localization and mapping, the semantic KITTI dataset~\cite{behley2019semantickitti} is a public dataset that provides annotated data for training and validation. Several navigation-related approaches have been proposed using semantic point clouds, such as semantic segmentation~\cite{milioto2019rangenet++}, semantic-aided LOAM\cite{li2021sa} and semantic Scan Context~\cite{li2021ssc}. Recent work Segregator~\cite{yin2023segregator} proposed to use semantic clusters and point features for global registration and achieves better performance compared to conventional geometric-only methods~\cite{yang2020teaser,lim2022single}.

This study is inspired by these previous semantic-aided works. The difference is that semantic-aided estimation is not trusted directly, and we propose to integrate geometric-level cues to verify the results. In addition, we test our proposed method in both indoor re-localization and outdoor loop closings, validating the extensibility of the proposed method.


\section{METHODOLOGY}
\label{sec:methods}

\begin{figure*}[t]
	\centering
	\includegraphics[width=\linewidth]{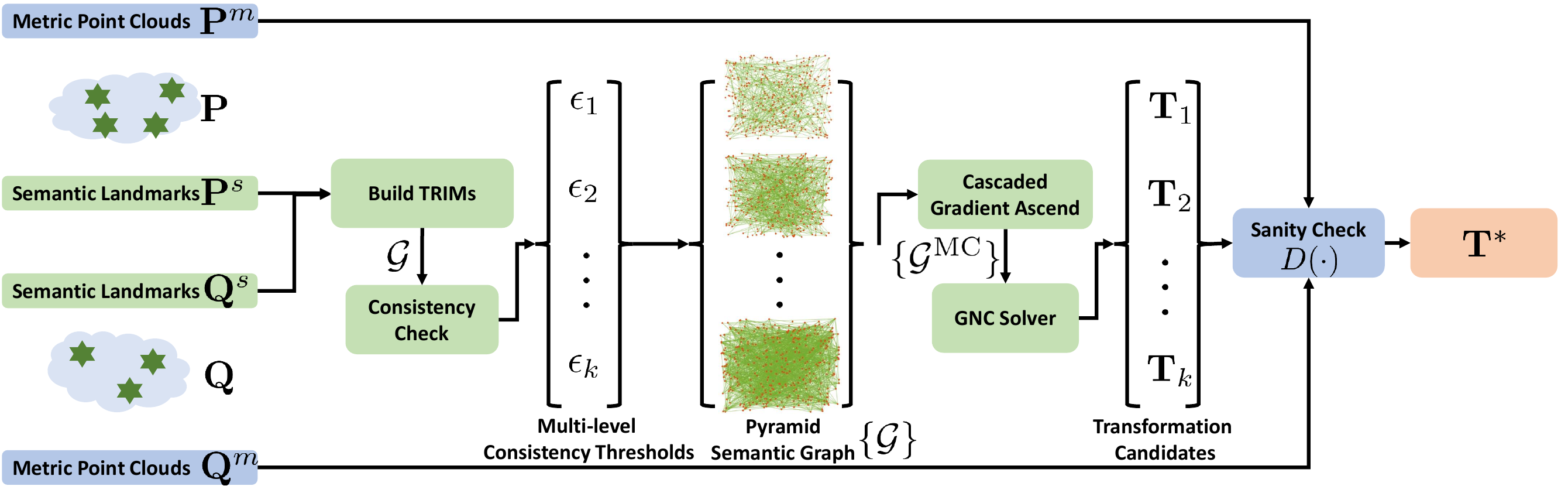}
	\caption{The proposed scheme follows a ``distrust and verify" strategy for global point cloud registration. It mainly consists of two parts: pyramid semantic graph-based registration (in green), and geometric verification using metric point clouds (in blue). The pyramid graph is built by shrinking the consistency threshold $\epsilon$, and it provides multiple transformation candidates for the following geometric verification. }
	\label{figure:scheme}
\end{figure*}

We first introduce the preliminary and problem formulation, then illustrate
the details of our proposed method, from front-end invariant measurements extraction to the final geometric verification. An overview of the proposed method is presented in Figure~\ref{figure:scheme}.

\subsection{Preliminary and problem formulation}

Given a source point cloud $\mathbf{P}=\{\mathbf{p}\}$ and target point cloud $\mathbf{Q}=\{\mathbf{q}\}$, we first generate semantic feature points (landmarks) $\mathbf{P}^s=\{\mathbf{p}^s\}$ and $\mathbf{Q}^s=\{\mathbf{q}^s\}$ using semantic cues, e.g., based on point cloud semantic segmentation, or visual-aided detection. Original metric point clouds $\mathbf{P}$ and $\mathbf{Q}$ are maintained as $\mathbf{P}^m$ and $\mathbf{Q}^m$. The aim of our approach is to estimate a transformation $\mathbf{T}=\left[\mathbf{R},\mathbf{t}	\right]$, in which $\mathbf{R}\in\text{SO(3)}$ and $\mathbf{t}\in\mathbb{R}^3$, as the global registration result.

As analyzed in the introduction section, there are three possible problems if dense point clouds $\mathbf{P}^m,\mathbf{Q}^m$ are used directly in graph theory-guided methods, especially when $\mathbf{P}^m$ and $\mathbf{Q}^m$ are with low overlap. In this study, we consider that a more reasonable way is to achieve global registration by first aligning sparse $\mathbf{P}^s$ and $\mathbf{Q}^s$, followed by geometric verification using dense $\mathbf{P}^m$ and $\mathbf{Q}^m$. We will introduce our detailed method in the following subsections.

\subsection{TRIMs and consistency check}

\begin{figure}[t]
	\centering
	\includegraphics[width=6cm]{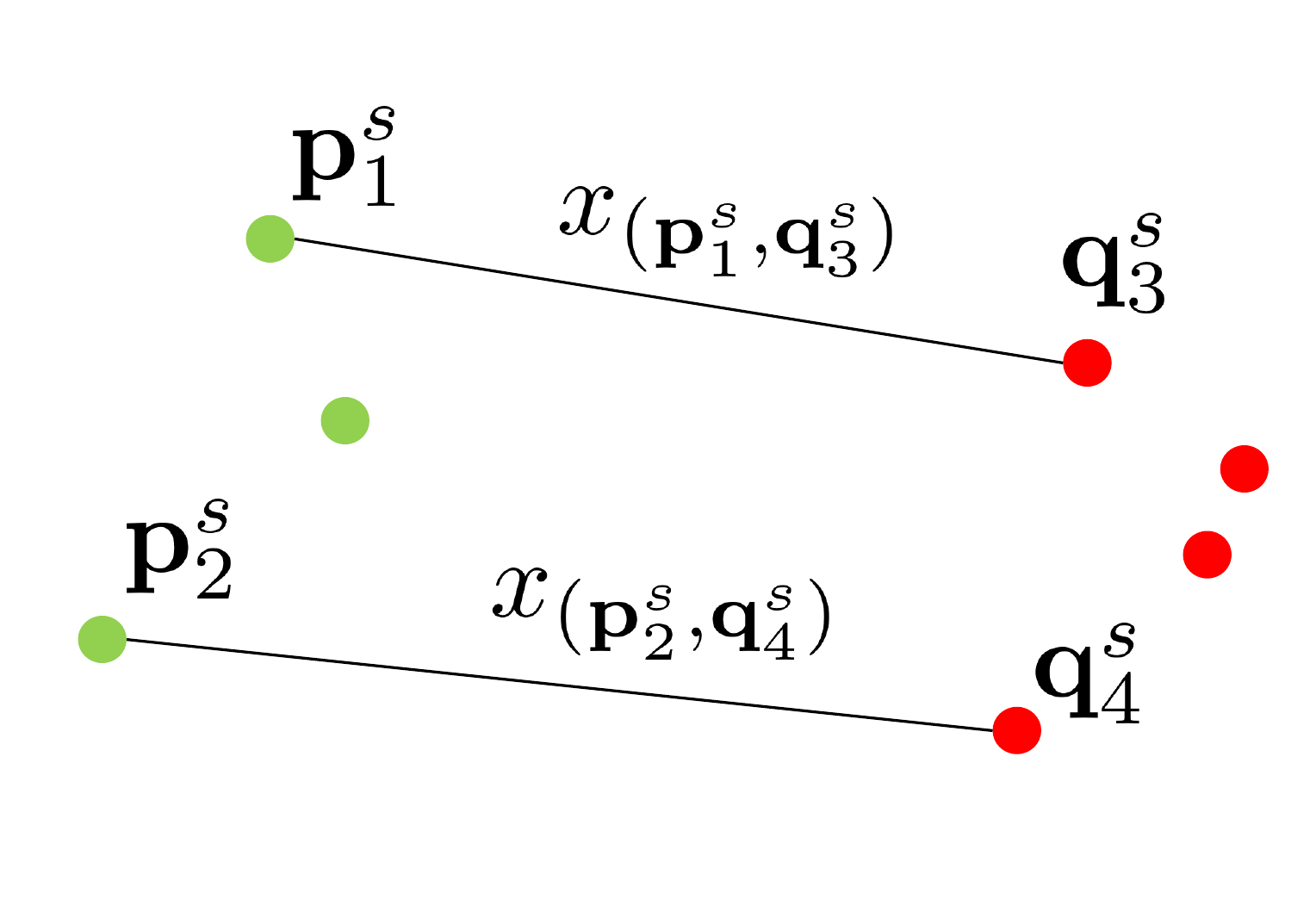}
	\caption{An example of two correspondences $(\mathbf{p}_1^s,\mathbf{q}_3^s)$ and $(\mathbf{p}_2^s,\mathbf{q}_4^s)$. Consistency threshold $\epsilon$ is used to check the consistency of two correspondences $x_{(\mathbf{p}_1^s,\mathbf{q}_3^s)}$ and $x_{(\mathbf{p}_2^s,\mathbf{q}_4^s)}$. }
	\label{figure:match}
\end{figure}

Optimally, $\mathbf{P}^s$ and $\mathbf{Q}^s$ are noise-free and provide fully correct correspondences for relative transformation estimation. However, the extracted semantic landmarks are not always accurate due to practical reasons, resulting in measurement noises followed by correspondence outliers. Most SoTA methods calculate translation and rotation invariant measurements (TRIMs) for outlier pruning. For example, by semantic-aided associations, two correspondences are built from $\mathbf{p}_1^s$ to $\mathbf{q}_3^s$, and from $\mathbf{p}_2^s$ to $\mathbf{q}_4^s$. The TRIM is defined as a difference $\Delta((\mathbf{p}_1^s,\mathbf{q}_3^s),(\mathbf{p}_2^s,\mathbf{q}_4^s)) :=  	\left| \left \| \mathbf{p}_1^s - \mathbf{p}_2^s \right\|	- \left \| \mathbf{q}_3^s - \mathbf{q}_4^s \right\| 	\right| $, or as a ratio $\left \| \mathbf{p}_1^s - \mathbf{p}_2^s \right\|	/ \left \| \mathbf{q}_3^s - \mathbf{q}_4^s \right\|$. We test these two kinds of definitions in the indoor (difference) and outdoor (ratio) experiments. Figure~\ref{figure:match} presents a visualized example for a better understanding of the TRIMs.

Then, a consistency threshold $\epsilon$ can be used to filter correspondence outliers based on TRIMs, written as follows

\begin{equation}\label{check}
\mathbf{W}((\mathbf{p}_1^s,\mathbf{q}_3^s),(\mathbf{p}_2^s,\mathbf{q}_4^s))  = \left\{\begin{array}{cc}
	1 ,& \text{if}~\Delta((\mathbf{p}_1^s,\mathbf{q}_3^s),(\mathbf{p}_2^s,\mathbf{q}_4^s)) \textless \epsilon \\
	0 ,& \text{otherwise}
\end{array}
\right.
\end{equation}
in which $\mathbf{W}(\cdot,\cdot)$ encodes the consistency check result for any two distinctive correspondences. Note that $\mathbf{W}(\cdot,\cdot)$ is binary in Equation~(\ref{check}), while it could be a weighted value if we assume a distribution on the noises, like Gaussian distribution-based~\cite{yin2023segregator}.

\subsection{Pyramid semantic graph}
\label{sec:PSG}

With the help of the consistency threshold $\epsilon$, we can build an affinity matrix $\mathbf{W}$ that describes the \textit{pairwise consistency} for all semantic-aided correspondences. $\mathbf{W}$ is a symmetric and binary/weighted matrix with a size of semantic correspondences, and the diagonal entries of $\mathbf{W}$ are set to $\mathbf{1}$ and the element that describe one-to-two matches are set to $\mathbf{0}$, e.g., $(\mathbf{p}_1^s,\mathbf{q}_3^s)$ to $(\mathbf{p}_1^s,\mathbf{q}_4^s)$. Essentially, $\mathbf{W}$ represents a consistency graph $\mathcal{G}$ that encodes all the semantic correspondences between $\mathbf{P}^s$ and $\mathbf{Q}^s$: the nodes are correspondences and the edges are pairwise consistencies.

Then the global registration problem is converted to estimate the maximum clique in a binary graph, or estimate the densest clique in a weighted graph~\cite{lusk2021clipper}. The estimated  graph $\mathcal{G}^\text{MC}$ should be a maximum fully connected subgraph $\mathcal{G}^\text{MC}\subseteq\mathcal{G}$. This problem can be formulated as Equation \ref{opt}, which is a typical NP-hard problem. Recent works proposed to approximate the optimal result by relaxation \cite{belachew2017solving,lusk2021clipper} or parallel search \cite{rossi2015parallel,yang2020teaser}.
\begin{equation} \label{opt}
\begin{aligned}
	& \underset{\mathbf{x} \in \left\lbrace 0,1 \right\rbrace^{\lvert\mathbf{W}\rvert}}{\text{maximize}}
	& & \frac{\mathbf{x}^\top \mathbf{W}\mathbf{x}}{\mathbf{x}^\top\mathbf{x}} \\
	& \text{subject to}
	& & x_ix_j=0 \quad \text{if} \  \mathbf{W}(i,j)=0, \forall_{i,j}
\end{aligned}
\end{equation}
where $\mathbf{x}$ is a binary vector that encodes the maximum clique or the densest clique: $x_i =1$ if correspondence is maintained in the clique, and vice versa. The estimated clique is regarded as an inlier set $\mathcal{I}$ for final transformation estimation. Finally, we solve a truncated least square problem with GNC~\cite{yang2020graduated}, written as follows
\begin{equation}
    \mathbf{T} = \left[ \mathbf{R},\mathbf{t} \right] = \argmin_{\mathbf{R}\in\text{SO(3)},\mathbf{t}\in\mathbb{R}^3} \sum_{ij\in\mathcal{I}} \text{min} \left( \left \| \mathbf{q}^s_i - \mathbf{R}\mathbf{p}_j^s - \mathbf{t} \right \|, c_{ij} \right)
\end{equation}
where $c_{ij}$ is the truncation parameter. GNC is robust to outliers in transformation estimation.

So far, our designed pipeline follows SoTA methods~\cite{yang2020teaser,lim2022single,yin2023segregator} and could be used to align semantic landmarks $\mathbf{P}^s$ and $\mathbf{Q}^s$ globally. But differently, we do not couple the rotation and translation for estimation, since semantic landmarks are sparser compared to feature points~\cite{rusu2009fast}, making one-step transformation easily calculated. Another critical difference is that we build an optimization formulation to achieve the solution graph $\mathcal{G}^\text{MC}$, different from the parallel search approaches in~\cite{yang2020teaser,lim2022single,yin2023segregator}. Optimization formulation could also help to solve the densest clique problem for weighted semantic-aided data. Our proposed cascaded gradient ascend will also benefit from this optimization.

As we analyzed in the introduction section, SoTA methods typically assume a noise bound around feature points or landmarks, and the consistency check is conducted once with a single threshold. We argue that there does not exist a perfect consistency threshold that can classify inliers and outliers in all cases. Thus, we propose to make multi-level thresholds, which could be a more reasonable choice for global registration.

To achieve this, we propose to loose the threshold from $\epsilon_{min}$ to $\epsilon_{max}$ to obtain $k$ thresholds:
\begin{equation} \label{eq:bound}
	\epsilon_{min}	\leq \epsilon_{i=1:k}	\leq	\epsilon_{max}
\end{equation}
in which $\epsilon_i$ should set according to a specific situation, as we mentioned in TRIMs design, and will be illustrated in the following experiments.

Thus with $k$ thresholds, there will be $k$ consistency graphs $\{\mathcal{G}\}$ and maximum/densest sub-graphs $\{\mathcal{G}^\text{MC}\}$, respectively. Theoretically, the larger $\epsilon$ is, the more pairwise correspondences will pass the consistency check, and the denser $\mathcal{G}$ and $\mathbf{W}$ will be. We rank all the consistency graphs $\mathcal{G}$ in according to their sparsity: sparsest at the top and densest at the bottom, shown in Figure~\ref{figure:scheme}. Our following cascaded gradient ascend-based solution will benefit from this kind of pyramid semantic graph.

\subsection{Cascaded gradient ascend}
\label{sec:cascaded}

If we solve the problems independently for each layer of the pyramid semantic graph, the time consumed will increase proportionally. In each layer, we follow \cite{lusk2021clipper} in relaxing the original problem to a continuous optimization problem with constraints.
\begin{equation}
\label{equ:mcp}
\begin{array}{cl}
\underset{\mathbf{x} \in \mathbb{R}_{+}^n}{\operatorname{maximize}} & F(\mathbf{x})=\mathbf{x}^{\top} \mathbf{W}_d \mathbf{x} \\
\text { subject to } & \|\mathbf{x}\| \leq 1
\end{array}
\end{equation}
where $\mathbf{x}$ denotes the probability of all correspondences, and
\begin{equation}
\mathbf{W}_d(i, j) \stackrel{\text { def }}{=} \begin{cases}\mathbf{W}(i, j) & \text { if } \mathbf{W}(i, j) \neq 0 \\ -d & \text { if } \mathbf{W}(i, j)=0\end{cases}
\end{equation}
where $d$ is a positive value and related to $\mathbf{W}_d$ and $\mathbf{x}$. (see \cite{lusk2021clipper} for more details). 

In this study, we propose a cascaded gradient ascend method to solve Equation~\ref{equ:mcp}. The main component of the proposed method relies on the projected gradient ascent algorithm (Projected GA) proposed in \cite{lusk2021clipper}. Still, it cannot be directly applied to solve the densest cliques of multiple consistency graphs. The core idea of our improvement is to solve the densest clique of each consistency graph in a cascaded way and then use the result obtained from the previous consistency graph as the initial value for the next iteration. 

\begin{algorithm}[t]
    \label{alg:cga}
    \caption{Cascaded Gradient Ascend}
    \SetKwInOut{Input}{Input}
    \SetKwInOut{Output}{Output}
    \Input{affinity matrices $\left\{\mathbf{W}_{1:k} \in [0,1]^{n\times n}\right\}$}
    \Output{densest cliques $\left\{\mathcal{G}^\text{MC}_{1:k}\right\}$ of $\left\{\mathcal{G}_{1:k}\right\}$}
    \BlankLine
    $\mathbf{x}_{0} \leftarrow \text{rand}(n,1)$\;
    \tcp*[h]{Default: consistency threshold of $\mathcal{G}_{i}$ is less than $\mathcal{G}_{i+1}$}\;
    \For{$i\leftarrow 1$ \KwTo $k$}{
            $\mathbf{x}_i, \mathbf{W}_d = \text{ProjectedGA}(\mathbf{W}_i, \mathbf{x}_{i-1})$\;
            $\hat{\omega}_i \leftarrow \text{round}(\mathbf{x}_{i}^{\top} \mathbf{W}_d \mathbf{x}_{i})$\;
            $\mathcal{G}^\text{MC}_{i} \leftarrow$ vertices corresponding to largest $\hat{\omega}_i$ elements of $\mathbf{x}_i$\;
        }
\end{algorithm}

Alg. \ref{alg:cga} explains the main flow of our algorithm. The process is straightforward, so here we only elaborate on two key designs of the proposed algorithm. First, we start from the sparsest graph to solve the maximum clique, which allows us to get a maximum clique solution as fast as possible. Second, since the set satisfying a low consistency threshold should be a subset of the one with a higher threshold, the maximum clique of the sparser graph tends to be contained within the one of the denser graph (note that although this is not absolute). In this case, the previous result can provide a better initial value, thus accelerating the projection gradient ascending.

\subsection{Geometric verification}
\label{sec:check}

We can obtain a set of transformation candidates $\{\mathbf{T}\}$ using the proposed pyramid semantic graph, and GNC solver. We then propose to pick the ``best'' transformation in the set $\{\mathbf{T}\}$ as the final result. To achieve this, the lidar scan $\mathbf{P}^m$ is first registered with a candidate $\mathbf{T}_i$ on the metric map $\mathbf{Q}^m$. Then we score each $\mathbf{T}_i$ with a truncated distance function between the two point clouds $\mathbf{T}_i\mathbf{P}^m$ and $\mathbf{Q}^m$, as follows:
\begin{equation}
D(\mathbf{T}_i\mathbf{P}^m, \mathbf{Q}^m) = \sum_{\mathbf{p} \in \mathbf{T}_i\mathbf{P}^m} \rho(\mathbf{p}, \mathbf{q}^\textbf{n})
\end{equation}
in which $\mathbf{q}^\textbf{n} \in \mathbf{Q}^m$ is the nearest map point for a point $\mathbf{p}$, and $\rho(\cdot)$ is a truncated Euclidean distance function:
\begin{equation} \label{eq:function}
\rho(\mathbf{p}, \mathbf{q}^\textbf{n})  = \left\{\begin{array}{cc}
	d ,&  d = \left \| \mathbf{p} - \mathbf{q}^\textbf{n} \right\| \textless \mu   \\
	\mu ,& d \geq \mu
\end{array}
\right.
\end{equation}
where $\mu$ is a distance threshold. The truncated function is more robust to the low overlapping $\mathbf{Q}^m$ and $\mathbf{P}^m$, and we will validate its effectiveness in the experiments.

Finally, the global localization result $\mathbf{T}^*$ can be estimated by selecting the minimum score $D(\cdot)$:
\begin{equation}
\mathbf{T}^* = \argmin_{i=1,\cdots,k} D(\mathbf{T}_i\mathbf{P}^m, \mathbf{Q}^m) 
\end{equation}

Intuitively, our proposed ``distrust and verify'' scheme is a combination of our designed graph-theoretic global registration (Section~\ref{sec:PSG} and Section~\ref{sec:cascaded}) and geometric-level verification (Section~\ref{sec:check}). 

From another point of view, it is also based on the assistance of two kinds of information: high-level semantics $\mathbf{P}^s$ and $\mathbf{Q}^s$ are noisy and we shrink the consistency thresholds to generate multiple hypotheses; then the low-level metric points $\mathbf{P}^m$ and $\mathbf{Q}^m$ can support the geometric verification to pick the most suitable hypothesis. The semantic cues and geometric cues play their own roles in this task, thus improving the performance of global point cloud registration, especially for challenging low overlapping point clouds.

\section{EXPERIMENTS}
\label{sec:experiments}

The proposed scheme is evaluated quantitatively in this section. We first introduce the experimental settings and then analyze the results compared to other state-of-the-art methods.

\subsection{Experimental settings}

We test the proposed method in both outdoor and indoor scenarios, and also with multi-session datasets, thus validating its effectiveness in an extensive way. The experimental settings are described as follows.

\subsubsection{Indoor re-localization} The proposed method is tested in a real-world university building for planar re-localization, i.e., $\mathbf{T}\in\text{SE(2)}$. We collect five sequences with a Velodyne VLP-16 lidar scanner for evaluation. These sequences are collected in two different stories and cover a large area of the floor plan. We manually set the start position of each sequence, and run a fine-tuned cartographer~\cite{hess2016real} to achieve all ground truth poses.

In each sparse laser scan, we tune the pole extractor~\cite{dong2021online} to extract columns as semantic landmarks. As for the map data, we sample 2D points on a floor plan map and set columns as semantics in the map. Semantic correspondences are built in an all-to-all manner from observed columns to columns on the map. Note that laser points and map points are with low overlaps since there exist deviations between as-observed data and the as-designed floor plan, making it very challenging to re-localize laser scans globally on the floor plan map.



\subsubsection{Outdoor loop closing} 
We also evaluate our proposed method on KITTI dataset for 3D loop closing, i.e., $\mathbf{T}\in\text{SE(3)}$. For a fair comparison with the SoTA method, we used the recent semantic global registration method, Segregator~\cite{yin2023segregator}, with the same experiment settings. Specifically, we tested on KITTI 00, 05, 06, 07, and 08, generating a large collection of loop-closing data for each sequence. According to the GT pose, the dataset is classified into three categories according to the translation between two point clouds: easy (3-5m), medium (8-10m), and hard (10-15m). Most methods have shown high success rates in the easy and medium datasets, as reported by~\cite{yin2023segregator}, so we only test on the hard dataset for evaluation in this experiment. The hard dataset contains point clouds with lower overlaps.

In addition, we choose the same semantic categories (car, truck, building) as Segregator for registration and use the dynamic curved-voxel clustering (DCVC)~\cite{zhou2021t} algorithm to segment the point clouds of each category into different sub-clouds. Different from Segregator, only centroids of sub-clouds are used as semantic landmarks. Since Segregator does not provide generated semantic labels in its open source code, we choose a fast but slightly weaker point cloud semantic segmentation network, Salsanext~\cite{cortinhal2020salsanext}, for semantic extraction. All-to-all strategy and FPFH-based matching are used to generate correspondences for cars (as well as trucks) and buildings, respectively.

\begin{table}[t]
	\renewcommand\arraystretch{1.0}
	\begin{center}
		\caption{Success Rate (\%) on Global Indoor Re-Localization}
		\label{table:recall}
  \resizebox{\linewidth}{!}{
		\begin{tabular}{p{3.2cm}<{\centering}|p{0.6cm}<{\centering}p{0.6cm}<{\centering}p{0.6cm}<{\centering}p{0.6cm}<{\centering}p{0.6cm}<{\centering}}
			\toprule
                \midrule
			Sequence & S1 & S2 & S3 & S4 & S5 \\
			Num. of Scans  & 1072 & 941 & 1737 & 1136 & 1572 \\
			\hline
         TEASER++ \cite{yang2020teaser} & NA & NA & NA & NA & NA \\
   Go-ICP~\cite{yang2015go} & 29.9 & 8.2 & 51.5& 20.6 & 24.6 \\
			RANSAC \cite{zhou2018open3d} (Sem)  & 27.8 & 15.8 & 42.3 & 36.5 & 35.0 \\
			TEASER++ \cite{yang2020teaser} (Sem) & 80.5 & 64.3 & 81.2 & 71.2 & 76.1 \\
			Ours & \textbf{86.4} & \textbf{80.7} & \textbf{96.8} & \textbf{77.6} & \textbf{83.9} \\
                \midrule
			\bottomrule
		\end{tabular}
  }
	\end{center}
\end{table}

\subsection{Evaluation of indoor re-localization}


\begin{table}[!t]
	\captionsetup{justification=centering}
	\renewcommand\arraystretch{1.0}
	\begin{center}
		\caption{Ablation Study on Geometric Verification on Sequence S1}
		\label{table:abla_distance}
        \resizebox{\linewidth}{!}{
		\begin{tabular}{p{0.6cm}<{\centering}|p{0.9cm}<{\centering}p{0.6cm}<{\centering}p{0.6cm}<{\centering}p{0.8cm}<{\centering}|p{1.5cm}<{\centering}p{0.7cm}<{\centering}}
			\hline
			\hline
			Eu &Ti-0.1 &Ti-1 &Ti-5 &Ti-10 &Tu-0.1(\cmark) &Tu-1 \\
			\hline
			 52.8 & 21.8& 50.6& 60.8& 39.8 & \textbf{76.4}& 76.2 \\
			\hline
			\hline
		\end{tabular}
        }
	\end{center}
 \vspace{-0.4cm}
\end{table}

\begin{figure}[!t]
	\centering
        \subfigure[Scene of Sequence S3]{	
		\includegraphics[width=4cm]{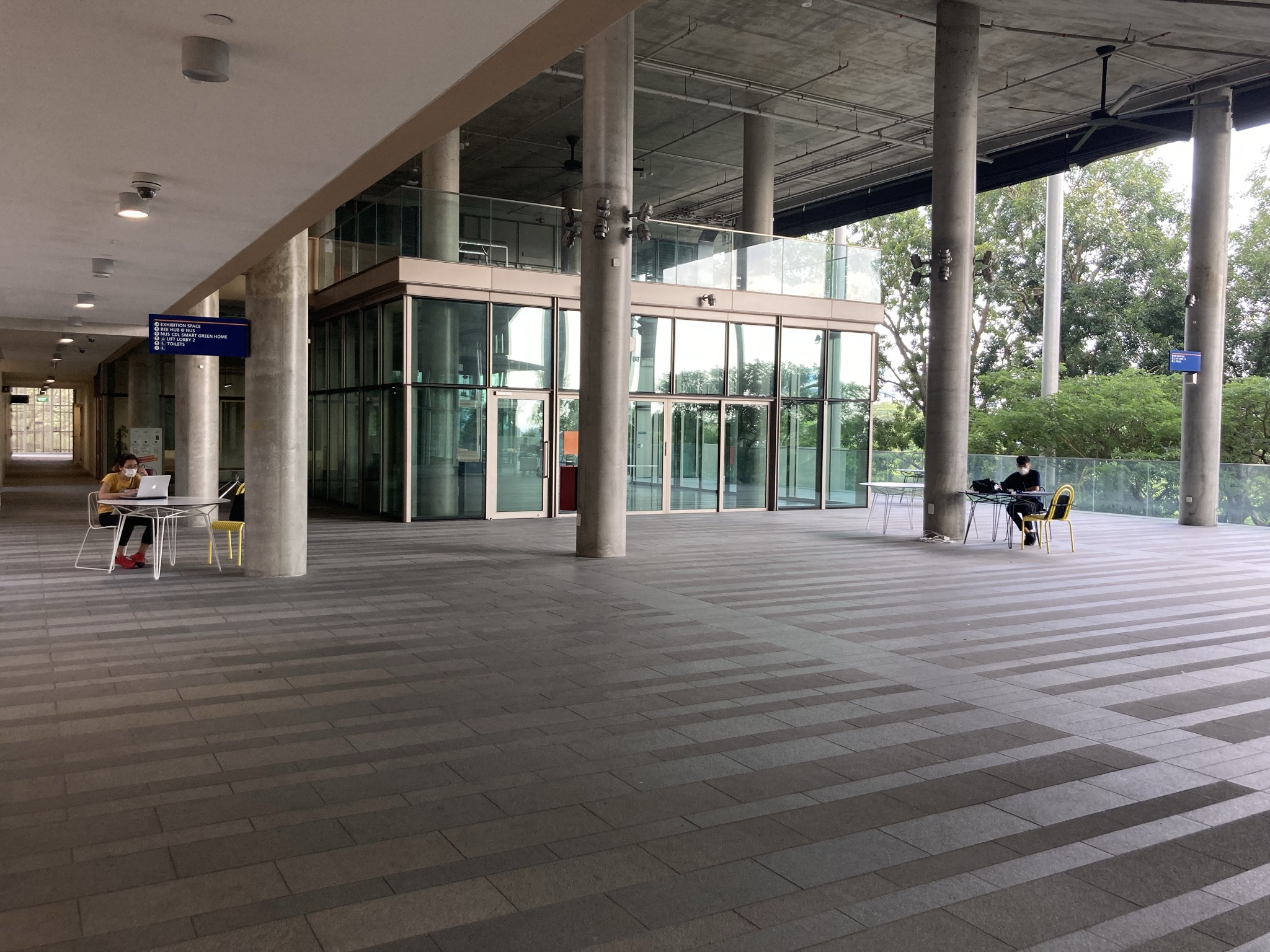}} 
	\subfigure[Scene of Sequence S5]{	
		\includegraphics[width=4cm]{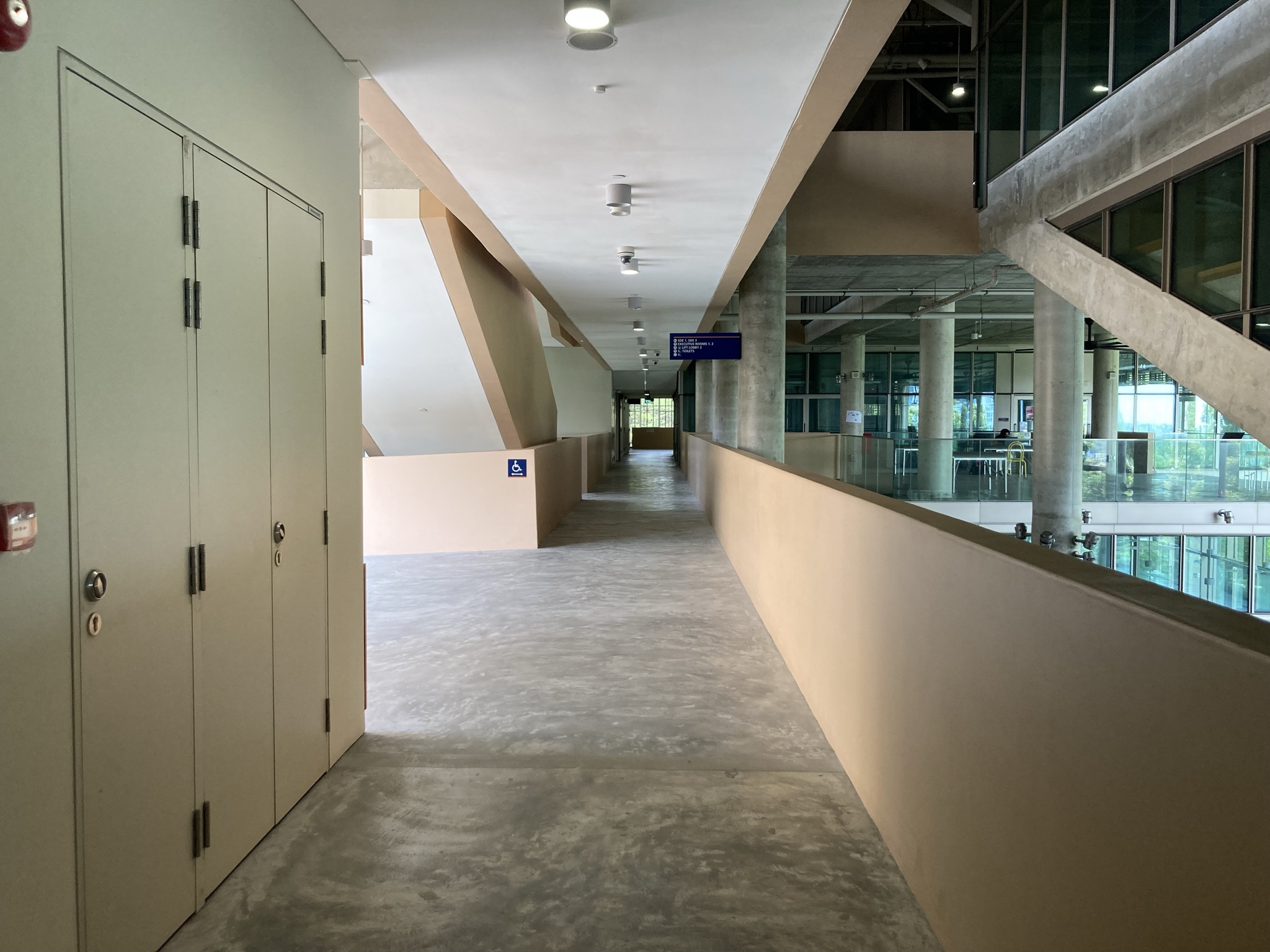}}
	\subfigure[Sequence S3 poses]{	
		\includegraphics[width=4cm]{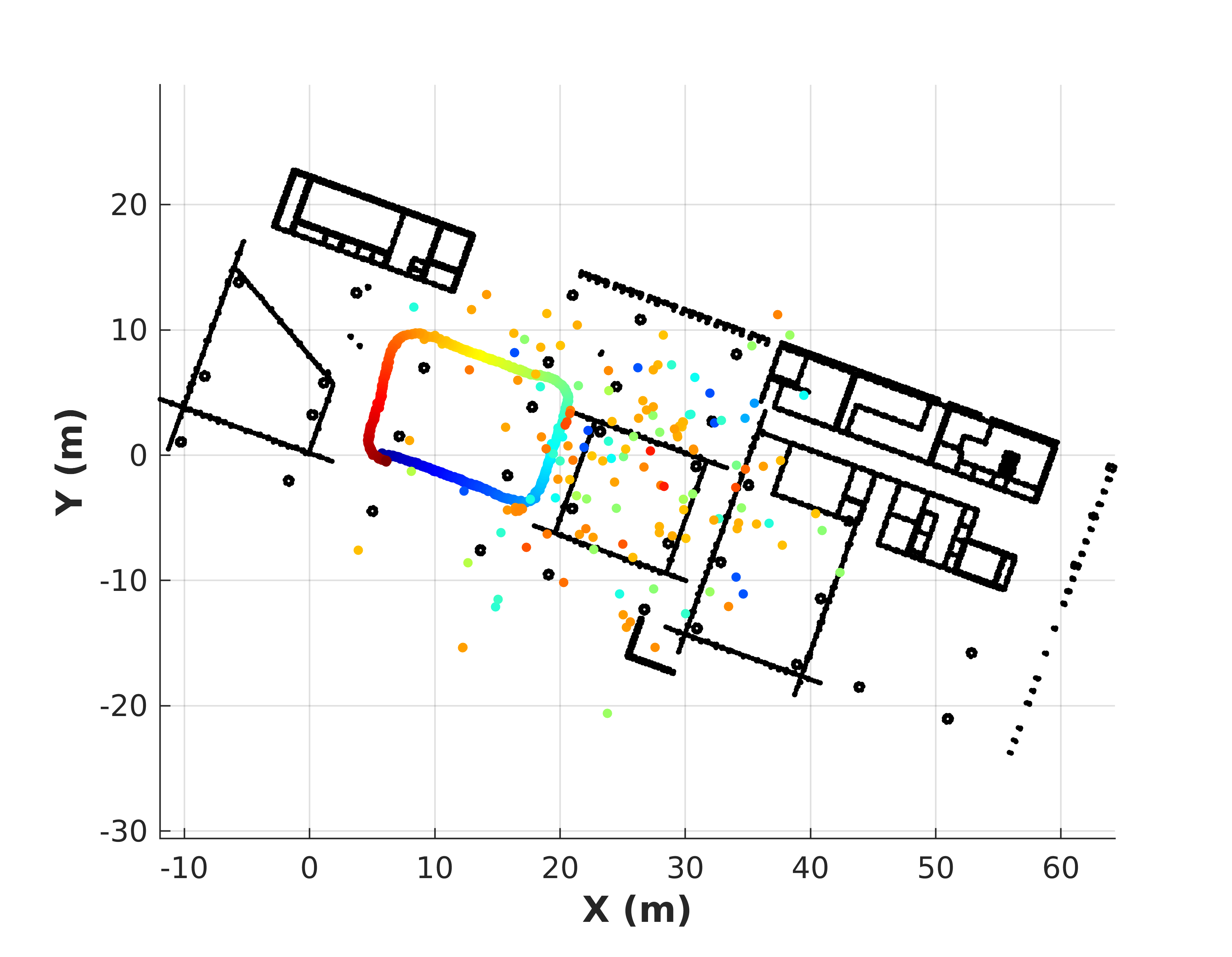}} 
	\subfigure[Sequence S5 poses]{	
		\includegraphics[width=4cm]{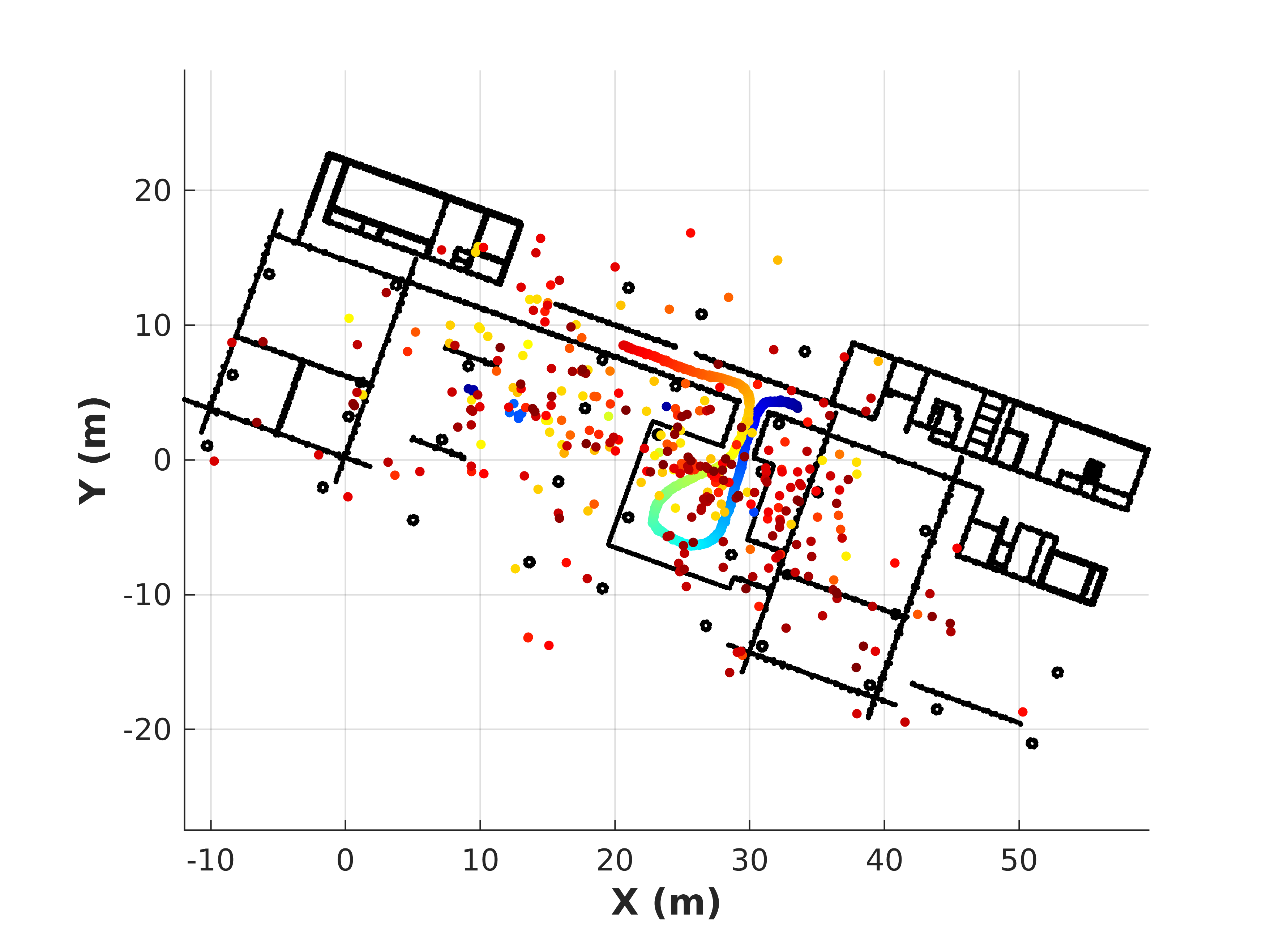}}
	\caption{(a)(b) Indoor scenes. (c)(d) The re-localized position at each timestamp. The estimated positions are colored from blue to red with respect to the time. These two sequences are collected at different stories.
        }
	\label{figure:trajectory}
\end{figure}

\begin{figure}[!t]
	\centering
	\includegraphics[width=8.5cm]{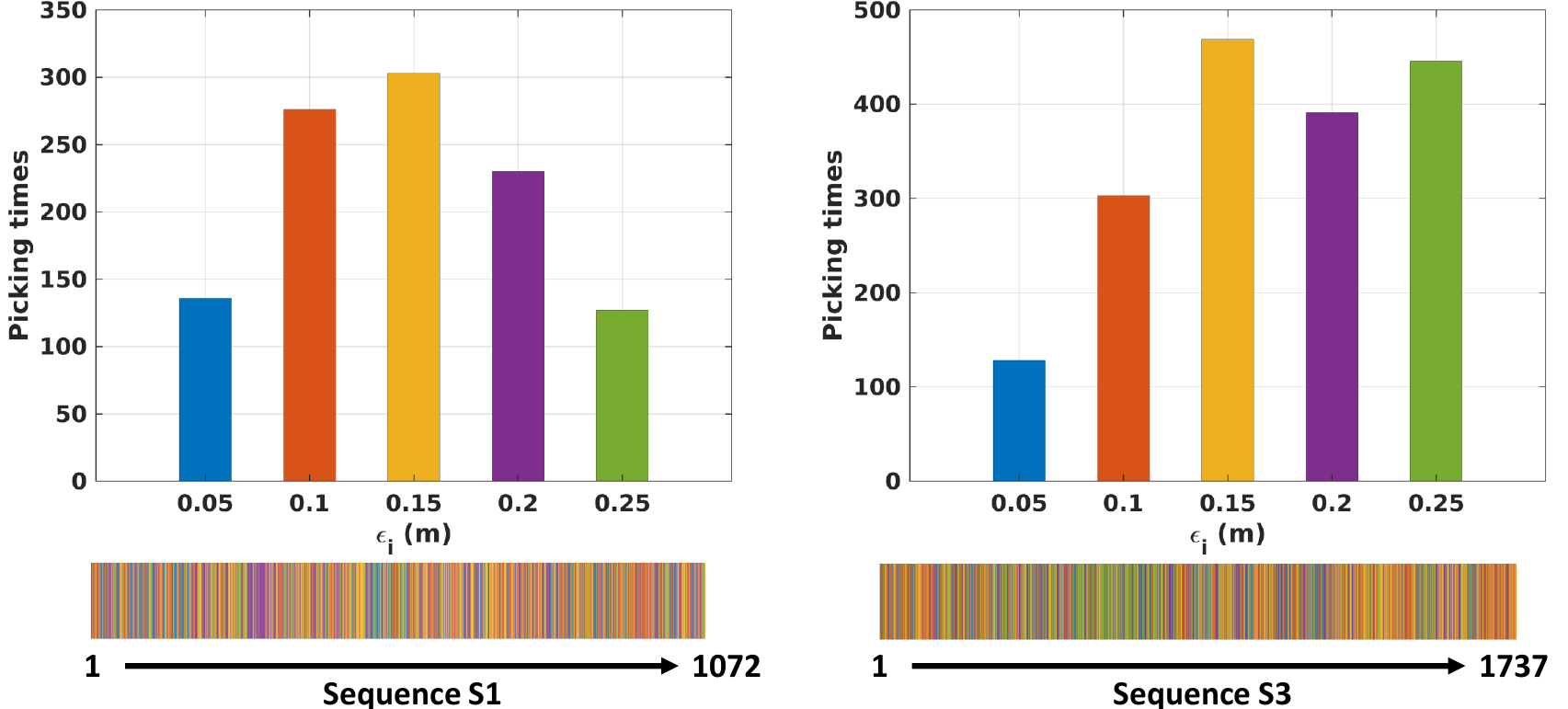}
	\caption{The upper figure shows the picking times of different threshold $\epsilon$, and the lower figure shows the ``best'' transformation verified by $\epsilon$  at each laser scan.}
	\label{figure:picking}
\end{figure}

\begin{figure*}[!t]
	\centering
	\includegraphics[width=18cm]{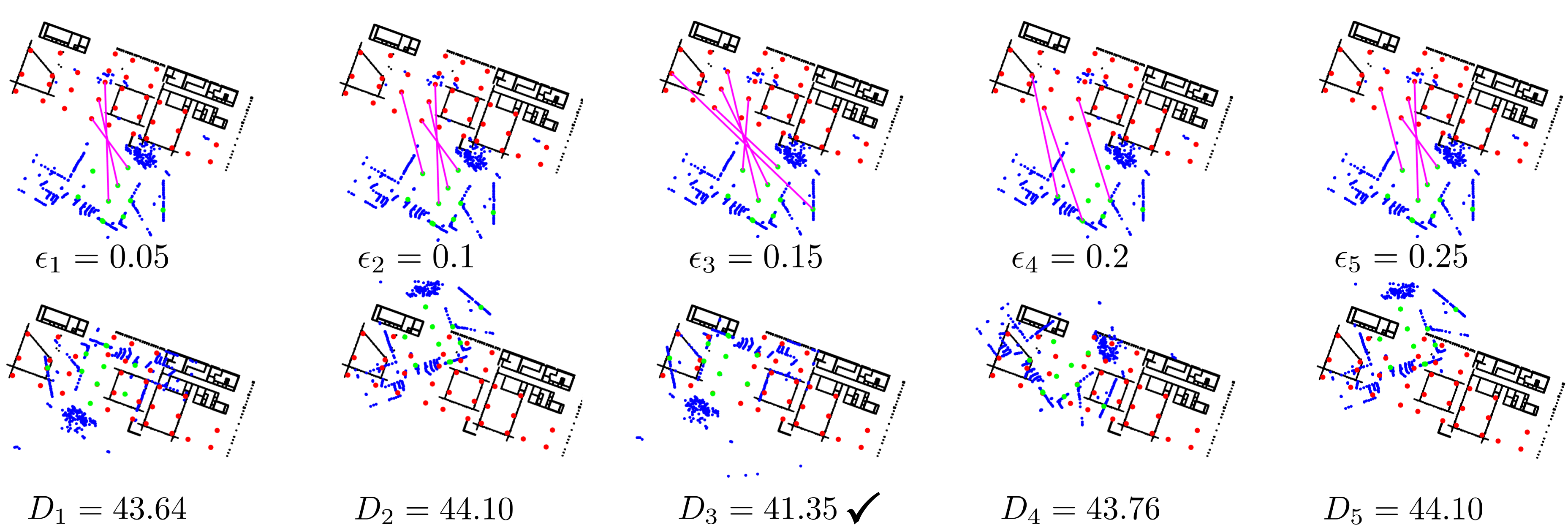}
	\caption{A case study on the 102nd scan on Sequence S1. The upper figures present the estimated correspondences with different thresholds $\epsilon$. Bold green and red points are $\mathbf{P}^s$ and $\mathbf{Q}^s$. Blue and black points are $\mathbf{P}^m$ and $\mathbf{Q}^m$, i.e, laser scan and floor plan map. The estimated maximum cliques are the same using $\epsilon_2$ and $\epsilon_5$. The lower figures present the global pose (transformation) estimation and the scores by geometric verification. Finally, $\mathbf{T}_3$ is verified as the final estimated global pose for the lowest score. The re-localized pose is correct (\textless 1m, \textless 3$^\circ$), though the overlap is very low between $\mathbf{T}_3\mathbf{P}^m$ and $\mathbf{Q}^m$.}
	\label{figure:case}
\end{figure*}

\subsubsection{Re-localization on map} We first evaluate the performance on the indoor re-localization task. Te proposed method is also compared to other competitive methods:
\begin{itemize}
\item We perform Go-ICP \cite{yang2015go} and TEASER++ (3D FPFH) \cite{yang2020teaser} based on the metric point map $\mathbf{Q}^m$. Go-ICP requires a large number of iterations for its branch-and-bound strategy. While feature points-based TEASER++ is inapplicable in this experiment, since it is extremely hard to build local feature-based associations \cite{rusu2009fast} using floor plan map.
\item For a fair comparison, we also perform RANSAC ~\cite{zhou2018open3d} (100K iterations) and TEASER++~\cite{yang2015go} based on all-to-all semantic landmark-based associations (column-to-column). Due to the lack of a pyramid graph, RANSAC and TEASER++ are tested without geometric verification on metric point clouds.
\end{itemize} 

We define a successful re-localization if the translational error is less than 1m and the rotational error is less than 3°. Table~\ref{table:recall} presents the evaluation results. Our proposed method performs better than the other methods, thus validating the effectiveness of using multi-level consistency thresholds for global point cloud registration. We also present the re-localized poses in Figure~\ref{figure:trajectory}.

Assuming multiple levels of consistency threshold is a key component in this study. We set the thresholds as 0.05m, 0.1m, 0.15m, 0.2m, 0.25m in the indoor experiments (column diameter is 0.6m). One might ask which is the best threshold for this re-localization experiment. We conduct analyses on the picking frequency of these thresholds in two sequences, shown in Figure~\ref{figure:picking}. All these five thresholds can contribute to global re-localization as a ``trusted'' threshold after geometric verification. Furthermore, a case study is presented in Figure~\ref{figure:case} to help understand the proposed distrust and verify strategy. 

\subsubsection{Ablation study} 

We also conduct an ablation study on the $\rho(\cdot)$ with different criteria and different values, which is the key to geometric verification. Specifically, the tested functions include a \textbf{T}r\textbf{u}ncated function (Tu) in Equation~\ref{eq:function}, a \textbf{T}r\textbf{i}mmed function ($\rho$=0 if $d \geq \mu$) (Ti) and the original nearest \textbf{Eu}clidean distance (Eu). Eventually, the trimmed function is close to a Huber weighted distance \cite{huber1992robust}, and the original nearest distances can be regarded as a reduced Chamfer metric \cite{barrow1977parametric}. In Table~\ref{table:abla_distance}, Ti-0.1 means that using the Trimmed function with $\mu$=0.1m. The results show that the truncated distance function performs better than the other two forms, which means it can pick more correct candidates.

\subsection{Evaluation of outdoor loop closing}

\subsubsection{Global registration}
We compare our methods with other three global point cloud registration methods, TEASER++~\cite{yang2020teaser}, Quatro~\cite{lim2022single}, and Segregator~\cite{yin2023segregator}, and a SoTA local registration method, V-GICP~\cite{koide2021voxelized}. For TEASER++, Quatro, and V-GICP, we use the evaluation results in the Segregator paper~\cite{yin2023segregator}. As with Segregator~\cite{yin2023segregator}, we define a successful registration if the translational error is less than 2m and the rotational error is less than 5°, and we count the success rate separately for each sequence. To obtain the pyramid graph, we prune the consistency using consistency thresholds of 0.01, 0.02, 0.04, and 0.08.

As shown in Table \ref{tab:sr_kitti}, the total number of tested point cloud pairs is up to 48026 frames using KITTI datasets. The results in Table~\ref{tab:sr_kitti} demonstrate the robustness of our proposed method on the low-overlap global registration task. It is worth noting that the performance of the Segregator is, to a larger extent, lower than the results of the original paper for some sequences. This is likely due to the slightly worse semantic segmentation performance of the Salsanext that we use. As discussed in the introduction section, poor data association and low overlap can cause the registration performance to be more sensitive to the consistency threshold. However, for our method, since we use multiple consistency thresholds for consistency graph construction, this largely enhances the robustness of the registration algorithm to incorrect semantics and low overlap. 

Finally, to check the performance of our proposed geometric verification, we give the upper bound of the proposed method in the last row of Table IV. Specifically, for the four generated transformation candidates, we determine whether one of them is successful by using the ground truth pose instead of using geometric verification. The results indicate that using a low-level geometric point cloud to validate the poses is reasonable.
\begin{table}[t]
	\caption{Success Rate (\%) on Global Registration \\with KITTI Hard Dataset (10-15m)}
        \vspace{-0.2cm}
	\label{tab:sr_kitti}
	\centering
	\resizebox{0.9\linewidth}{!}{
        \begin{tabular}{cccccc}
        	\toprule
         \midrule
        	\multicolumn{1}{c}{Sequence} & \multicolumn{1}{c}{00} & \multicolumn{1}{c}{05} 
            & \multicolumn{1}{c}{06} & \multicolumn{1}{c}{07} & \multicolumn{1}{c}{08}  \\
        	Num. of Pairs                      & 21267 & 11600 & 2825 & 2877 & 9457 \\ \midrule
        	V-GICP\cite{koide2021voxelized}                      & 14.9 & 25.3 & 35.1 & 0.80 & 4.90 \\
        	TEASER++\cite{yang2020teaser}                      & 41.1 & 48.0 & 83.2 & 30.7 & 49.5 \\
        	Quatro\cite{lim2022single}                      & 41.2 & 48.1 & 83.2 & 31.9 & 49.8 \\
        	Segregator\cite{yin2023segregator}              & 79.70 & 62.82 & 86.90 & 76.54 & 58.62 \\
        	Ours                            & \bf{90.38} & \bf{89.36} & \bf{97.42} & \bf{88.43} & \bf{81.66} \\ 
        	Ours (Upper)                            & 90.53 & 89.65 & 97.45 & 88.56 & 81.89 \\ \midrule
        	\bottomrule
    	\end{tabular}
    }
	\vspace{-0.4cm}
\end{table}
\subsubsection{Ablation Study}
We first conduct an ablation study on different consistency thresholds and present the success rate and time cost. As shown in Table \ref{tab:ablation_kitti}, different threshold sets are tested with the same experimental setting. We start with the smallest consistency threshold and add a larger threshold one by one to generate a denser consistency graph. We also combine different thresholds for evaluation. Table \ref{tab:ablation_kitti} indicates that a larger consistency threshold can help improve performance. However, we also notice that in some cases, when we add a too-large threshold, it helps a little to improve the registration performance. We consider that when the consistency threshold is increased to a certain value, more outliers can pass the consistency check, making a consistency graph too large. Thus the ratio of correspondence inliers ($\mathcal{G}^\text{MC}$) is very small in the too-large graph ($\mathcal{G}$), confusing the optimization-based densest clique solver. Table \ref{tab:ablation_kitti} also indicates that the success rate increases when we add a threshold into the set. The aggregated threshold set generally outperforms any individual threshold subset. However, it plays a minimal role when the number of thresholds is increased to a certain level. It may be because when introducing more thresholds, the estimated cliques could be the same for these consistency graphs, resulting in similar performances for four-threshold sets and five-threshold sets.


Table \ref{tab:time_kitti} presents the runtime costs to solve the densest clique varies as the number of consistency thresholds increases from one to five. For comparison, we also give the time variation of the robust transformation estimation (GNC solver). The values in parentheses represent the multiplication of time cost with respect to one consistency threshold (second column). It can be seen that the runtime of the GNC solver is almost proportional to or slightly larger than the number of thresholds because each estimation is independent of the other. Larger thresholds lead to denser graphs, resulting in more correspondences and more solving time in denser graphs. On the other hand, for the densest clique solver, the cascaded gradient ascending algorithm can re-use a previous result. The re-used results are transferred to the denser as an initial guess, which is better than the random guess for the optimization problem. Table \ref{tab:time_kitti} demonstrates that our proposed cascaded gradient ascending algorithm can speed up the solving process in the pyramid semantic graph. 


\begin{table}[t]
	\caption{Ablation Study on Different Consistency Thresholds \\ Success Rate (\%)}
	\vspace{-0.2cm}
	\label{tab:ablation_kitti}
	\centering
	\resizebox{0.9\linewidth}{!}{
        \begin{tabular}{cccccc}
        	\toprule
         \midrule
        	\multicolumn{1}{c}{Sequence} & \multicolumn{1}{c}{00} & \multicolumn{1}{c}{05} 
            & \multicolumn{1}{c}{06} & \multicolumn{1}{c}{07} & \multicolumn{1}{c}{08}  \\ \midrule
        	$[0.01]$                           & 84.13 & 78.22 & 96.00 & 82.10 & 72.00 \\
        	$[0.02]$                           & 86.03 & 84.96 & 96.46 & 83.73 & 77.47 \\
        	$[0.01, 0.02]$                     & 87.58 & 83.47 & 96.60 & 85.57 & 77.03 \\
        	$[0.04]$                           & 84.69 & 88.97 & 96.92 & 82.55 & 78.79 \\
        	$[0.01, 0.02, 0.04]$               & 89.69 & 87.34 & 97.20 & 87.42 & 80.06 \\
        	$[0.08]$                           & 82.17 & 86.81 & \bf{97.49} & 79.77 & 78.67 \\
        	$[0.01, 0.02, 0.04, 0.08]$         & \bf{90.38} & \bf{89.36} & \bf{97.42} & \bf{88.43} & \bf{81.66} \\ \midrule
        	\bottomrule
    	\end{tabular}
    }
\end{table}

\begin{table}[t]
	\caption{Runtime (ms) Analysis on Different Consistency Thresholds}
	\vspace{-0.2cm}
	\label{tab:time_kitti}
	\centering
	\resizebox{\linewidth}{!}{
        \begin{tabular}{cccccc}
        	\toprule
         \midrule
        	\multicolumn{1}{c}{Num. of Thresholds} & \multicolumn{1}{c}{1} & \multicolumn{1}{c}{2} 
            & \multicolumn{1}{c}{3} & \multicolumn{1}{c}{4} & \multicolumn{1}{c}{5}  \\ \midrule
        	Clique Solver      &  2.98 (1) & 3.78 (\bf{1.27}) & 5.65 (\bf{1.89}) & 9.47 (\bf{3.18}) & 14.14 (\bf{4.74}) \\ \midrule
        	GNC Solver   & 1.79 (1) & 3.54 (1.98) & 5.63 (3.15) & 7.74 (4.32) & 10.54 (5.89) \\ \midrule
        	\bottomrule
    	\end{tabular}
    }
    \vspace{-0.4cm}
\end{table}


\section{CONCLUSIONS}

In this paper, we propose a novel semantic-geometric graph-theoretic method, aiming at registering point clouds with low overlap. The entire framework follows a ``distrust and verify'' strategy. The research insight is building a pyramid semantic graph with multiple consistency thresholds, which can provide multiple transformation candidates. A geometric-level verification is also designed for final estimation based on aligned low overlapping point clouds.

The proposed method is extensively validated using challenging indoor re-localization datasets and outdoor KITTI datasets. The results demonstrate that our method could achieve global registration with a high success rate, though the overlap between point clouds is low. Our future work includes integrating this method into Monte Carlo localization, and also includes building a robust map merging model for city-scale mapping. In addition, more theoretical analyses are desired to improve the performances in the future study.

\bibliographystyle{IEEEtran}
\bibliography{root}

\end{document}